\documentclass[letterpaper, 10 pt, conference]{IEEEtran}
\def\IEEEtitletopspace{18pt}
\IEEEoverridecommandlockouts
\usepackage[nolist]{acronym}
\usepackage{amsmath,amssymb,amsfonts}
\usepackage{algorithmic}
\usepackage{graphicx}
\usepackage{textcomp}
\usepackage{xcolor}
\usepackage{svg}
\usepackage{adjustbox}
\usepackage{pgfplots}
\usepgfplotslibrary{statistics}
\usepackage{pgfplotstable}
\usepackage{booktabs}
\usepackage{multirow}
\usepackage{float}
\usepackage{multicol}
\usepackage{todonotes}
\usepackage{caption}
\usepackage{stfloats}
\usepackage{subcaption}
\usepackage{siunitx}
\usepackage{hyperref}
\usepackage{pifont}
\usepackage{placeins}
\usepackage[capitalise]{cleveref}
\newlength{\IEEEleft}
\usepackage[absolute,overlay]{textpos}
\def\BibTeX{{\rm B\kern-.05em{\sc i\kern-.025em b}\kern-.08em
    T\kern-.1667em\lower.7ex\hbox{E}\kern-.125emX}}

\usepackage{hyperref}
\usepackage{overpic}
\hypersetup{
    colorlinks=true,
    linkcolor=black,
    citecolor=black,
    filecolor=black,
    urlcolor=black,
}

\usepackage[style=ieee,
            doi=false,
            url=false,
            mincitenames=1,
            maxcitenames=1,
            minbibnames=6,
            maxbibnames=6,
            backend=biber]{biblatex}  % supports bibliography with BibLaTeX
\AtBeginBibliography{\footnotesize}

\begin{acronym}
	\acro{GNSS/INS}{global navigation satellite system/inertial navigation system}
\end{acronym}

\renewcommand{\tablename}{Table}
\makeatletter
\renewcommand{\@maketitle}{%
  \newpage
  \null
  \vskip 2em%
  \begin{center}%
  \let \footnote \thanks
    {\huge \@title \par}% Changed from \Huge to \Large
    \vskip 1.5em%
    {\large
      \lineskip .5em%
      \begin{tabular}[t]{c}%
        \@author
      \end{tabular}\par}%
    \vskip 1em%
  \end{center}%
  \par
  \vskip 1.5em}
\makeatother

\begin{document}

\title{
    Radar-based Pose Optimization for HD Map Generation from Noisy Multi-Drive Vehicle Fleet Data
}

\author{
    Alexander Blumberg$^{1}$,
    Jonas Merkert$^{1}$ and 
    Christoph Stiller$^{1}$ % <-this % stops a space
    \thanks{
        $^{1}$Institute of Measurement and Control Systems, Karlsruhe Institute of Technology (KIT),
        Karlsruhe, Germany
        {\tt\footnotesize \{alexander.blumberg, jonas.merkert, stiller\}@kit.edu}
    }%
}%

\maketitle

\begin{textblock*}{\textwidth}(\IEEEleft, 265mm) % adjust Y only if needed
\footnotesize
© 2026 IEEE. Personal use of this material is permitted. Permission from IEEE must be obtained for all other uses, in any current or future media, including reprinting/republishing this material for advertising or promotional purposes, creating new collective works, for resale or redistribution to servers or lists, or reuse of any copyrighted component of this work in other works.
\end{textblock*}

\begin{abstract}
High-definition (HD) maps are important for autonomous driving, but their manual generation and maintenance is very expensive. This motivates the usage of an automated map generation pipeline. Fleet vehicles provide sufficient sensors for map generation, but their measurements are less precise, introducing noise into the mapping pipeline. 

This work focuses on mitigating the localization noise component through aligning radar measurements in terms of raw radar point clouds of vehicle poses of different drives and performing pose graph optimization to produce a globally optimized solution between all drives present in the dataset.

Improved poses are first used to generate a global radar occupancy map, aimed to facilitate precise on-vehicle localization. Through qualitative analysis we show contrast-rich  feature clarity, focusing on omnipresent guardrail posts as the main feature type observable in the map.

Second, the improved poses can be used as a basis for an existing lane boundary map generation pipeline, majorly improving map output compared to its original pure line detection based optimization approach. 
\end{abstract}

\section{Introduction}

High-definition (HD) maps are the basis of modern autonomous vehicles. Contrary to standard-definition (SD) navigation maps, HD maps aim to provide more detailed, more precise and additional information about the relations between road elements. Its applications inside an autonomous vehicle stack range from localization inside the map to map-based planning to map-based object prediction. However, the creation of such maps is non-trivial  and mainly done using mobile mapping platforms equipped with highly precise sensors and reference global localization systems. Constant changes in road layout as a result of construction limit the validity of such maps and require extensive maintenance.

\begin{figure}[h]
    \centering
    \begin{overpic}[width=\columnwidth, trim=15cm 2cm 15cm 2cm, clip]{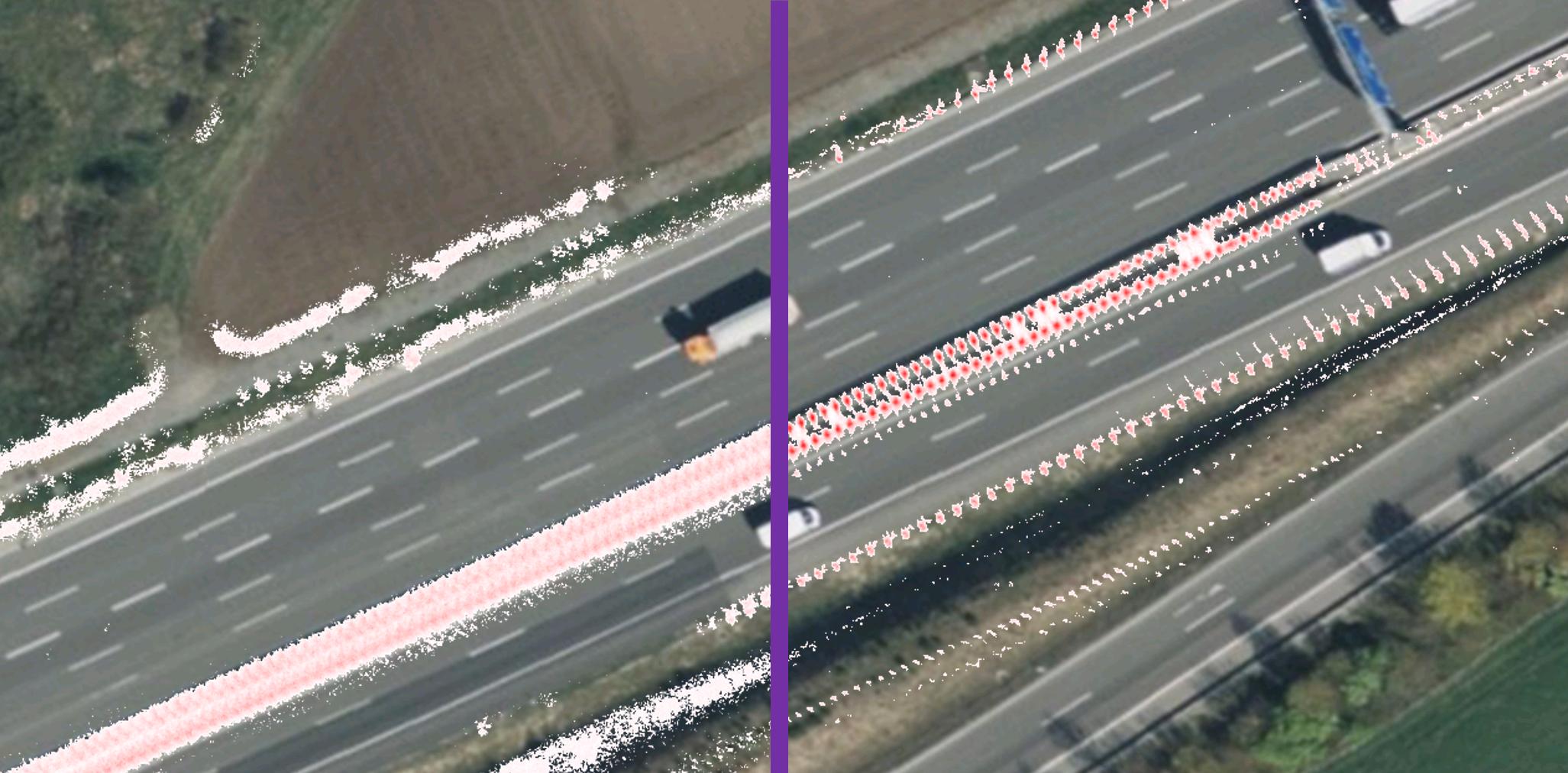}
        \put(0,0.5){\color{white}\scalebox{.4}{Aerial images from Bing Maps \textcopyright  2025 Microsoft}}
    \end{overpic}
    \caption{Radar occupancy map over satellite image: Left with unaligned, right with aligned vehicle poses. The equally spaced posts of the middle guardrail section become clearly distinguishable after alignment.}
    \label{fig:heatmap}
\end{figure}

Modern autonomous driving level 2 or 3 enabled customer vehicles are, however, already equipped with a broad range of sensors including cameras and radar sensors. Therefore they often are able to already provide road feature predictions of certain quality. While novel machine learning based approaches aim to facilitate solving the problem through online HD map construction ~\cite{liao2023MapTR, liao2024maptrv2, liu2023vectormapnet, chen_maptracker_2025, wang2024StreamQueryDenoising}, these proposals do not reach the levels of offline generated HD maps yet.

To still utilize the vast potential a vehicle fleet offers regarding HD map generation, one could upload all sensor data, including camera sensor data, for combined offline fusion and processing. However, this would require vast networking resources. An approach that limits this necessity for resources could therefore aim to only upload less size intensive sensor data like radar point-clouds as well as road feature detections already generated by the vehicle itself. Since both sensor measurements as well as ego localization of such vehicles are less reliable, further processing of this data in terms of alignment between vehicle poses is necessary \cite{immel2023hd}.

The main contributions of this work are:
\begin{itemize}
    \item We propose a robust approach for obtaining aligned poses based on radar point clouds from vehicle fleet data, exceeding existing methods for our particular data.
    \item We aggregate and fuse these radar point clouds and provide a radar occupancy map.
    \item We examine the improvements possible through utilizing pure radar based alignment compared to pure detected lane marking and road boundary based alignment in terms of generated HD lane boundary map quality.
\end{itemize}

\section{Related work}
Generating maps from fleet data or crowdsourcing is a longstanding research interest.

Whereas a major portion of the work focuses on SD map generation~\cite{10.1109/IVS.2019.8813860}, \cite{GPS_Community_Map}, some research has been done on multi-journey data for HD map generation. Based on recorded video data, \cite{8968020} proposes to detect road feature elements in single drives and based on matching features perform pose graph optimization between drives.

\cite{DoerHenzlerMessner2020_1000126763} and \cite{dabeer2017end} work on fusing already generated lane marking and road boundary detections and perform fleet data mapping formulated as a Graph SLAM.

Previous works on radar based localization employ Iterative Closest Point  (ICP) for alignment between radar point clouds~\cite{7535489Ward}. Rai-slam~\cite{doppler} additionally includes Doppler velocities during radar scan matching and employs both local and global pose graph optimization to provide a SLAM approach. \cite{landmark} employs Extended Kalman Filter
(EKF) and Rao-Blackwellised Particle Filter (RBPF) implementations for a landmark-based radar Simultaneous Localization and Mapping (SLAM).

While previous works address mapping at the SLAM level, focusing on the single drive, the focus of this work specifically lies in overcoming the GNSS localization error with fleet data and an optimization across multiple drives.

\subsection{Fleet data map generation}
\label{subsec:sd4mad}
Immel et al.~\cite{immel2023hd} introduce a complete HD map generation pipeline based on camera based road feature detection. This provides both a baseline as well as lane boundary map generation and evaluation tooling for this work. This subsection reviews parts of this pipeline.

The approach is based on vehicle fleet data containing trajectories as well as high-level road features like lane markings and road boundaries. Stepwise lateral sampling of detections is performed. Detections of the same class are clustered and aligned using Expectation Maximization (EM).

Based on the maxima of the probability density function of the aligned clustered points, a polyline graph can be generated between maxima of consecutive steps, resulting in a lane-divider based map representation.

\begin{figure}[h]
    \centering
    \adjustbox{trim=0cm 0cm 0cm 0cm, width=\columnwidth, clip}{
        \includesvg[width=\columnwidth]{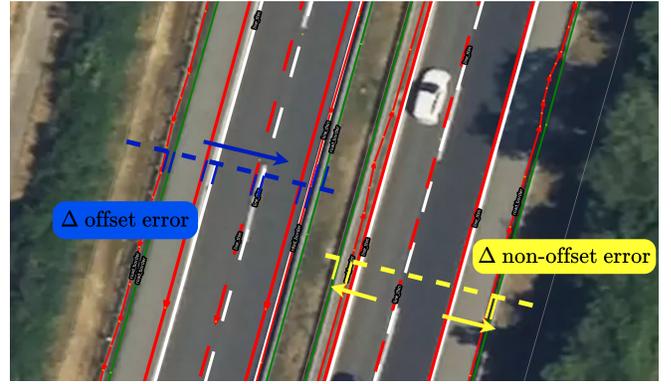}
    }
    \caption{Comparison of evaluation metrics: mapping pipeline output (red), ground truth road boundary (green), and lane dividers (white, solid and dashed). The offset error is visualized in blue, the remaining error after offset correction, therefore the non-offset error, in yellow.}
    \label{fig:errors}
\end{figure}

Furthermore, means of evaluation against ground truth are also introduced. This similarly includes stepwise lateral sampling and is based on the same association logic as the optimization step. Based on this association a lateral error can be calculated. This error consists of two parts. First, the lateral offset error, which is caused by the remaining localization error after optimization as well as improperly georeferenced ground-truth. Second, after accounting for the offset error, an uncorrectable error remains, the non-offset error as inconsistent spacing between map elements of the generated map and ground truth. This spacing may be caused by both limited detection accuracy by the vehicle as well as by inaccurately labeled/non-matching ground-truth. \cref{fig:errors} shows both errors.

\FloatBarrier
\begin{figure*}[h]
    \centering
    \includegraphics[width=\textwidth]{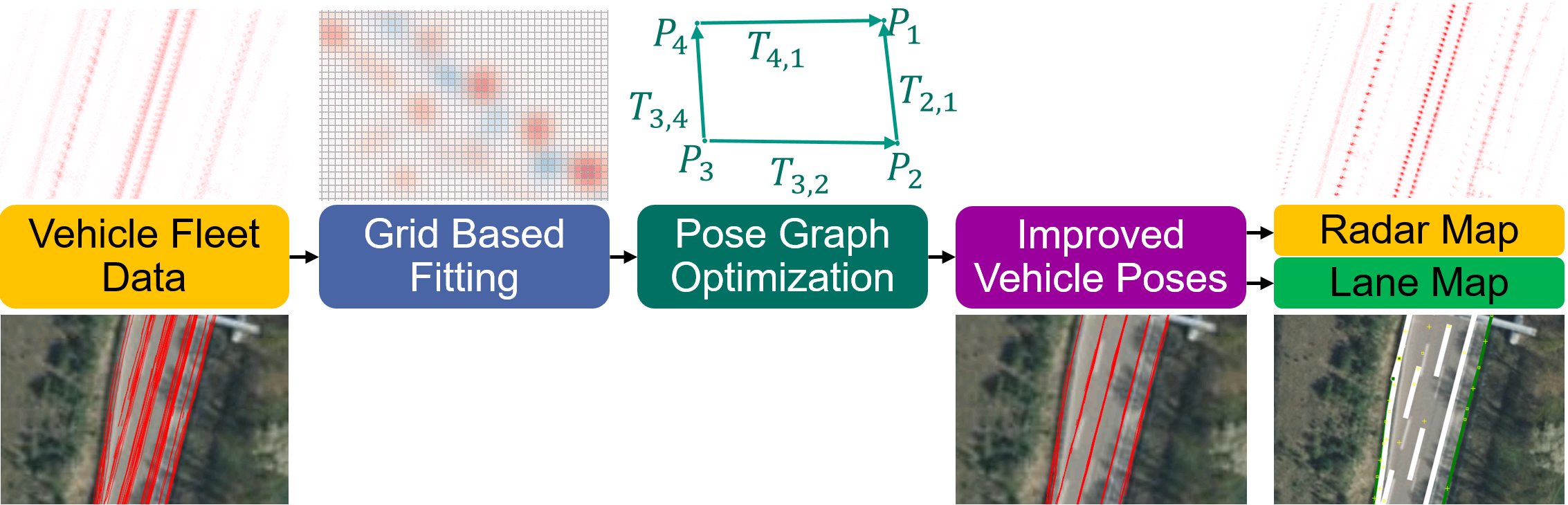}
    \caption{Flowchart showing all components of the map generation process. The dataset is described in \cref{subsec:dataset}, correlation calculation based on grid-based fitting in \cref{subsec:correlation}, the pose graph optimization in \cref{subsec:posegraph}, the generation of the occupancy map in \cref{subsec:occu}, and of the lane marking based map in \cref{subsec:lane}.}
    \label{fig:overview}
\end{figure*}

\section{Methodology}
\label{sec:meth}
\cref{fig:overview} provides an overview of the pipeline being introduced in this work. The following chapter will introduce all its modules in more detail.
\subsection{Input Data}
\label{subsec:data}
The algorithm aims to process a higher number of drives from a vehicle fleet with locally overlapping routes. Although drives may belong to different routes on the highway network, a sufficient amount of overlap is necessary in order to make the algorithm a viable option. Besides GNSS based trajectory information, drives include 2D 360° radar point clouds as well as road feature detections in the form of classified polylines for every pose. Since timestamps between trajectories and radar/feature frames do not always match, trajectories are interpolated using cubic Hermite spline for both position and heading.  The dataset is introduced in \ref{subsec:dataset} in more detail.

This work aims to overcome the limited accuracy of the GNSS poses provided by aligning neighboring poses between drives and within drives to each other and performing pose-graph optimization, aiming to overcome the Gaussian-distributed error of the GNSS poses.

\subsection{Data sampling}
\label{subsec:data_sampling}
To limit computational requirements in calculating correlations between vehicle poses both of the same drive as well as different drives, a list of "what-to-compare" is generated in advance. 

Candidates consist of pairs of vehicle poses and are filtered according to their geographical distance not exceeding 20m and being of different drives. Since a further rule-based selection of candidates is non-trivial, a simple random sampling is employed. For the data at hand, a sampling rate of 10 \% is used, which proved sufficient for the data at hand. These candidates are then registered in the "what-to-compare" list. Additionally we include consecutive poses as pairs in our "what-to-compare" list as well.

\subsection{Correlation}
\label{subsec:correlation}
Based on the previously defined list of "what-to-compare", we go through each pair within this list individually and perform the correlation calculations for each pair. For one pair of 6D vehicle poses $\mathbf{x}_1$ and $\mathbf{x}_2$ this represents calculating the transformation matrix $T_{1,2}$.

We compare different methods for correlation calculation as follows:

\subsubsection{Baseline}
To provide a comparative baseline, algorithms as implemented by Fast-GICP~\cite{EasyChair:2703} are utilized. More specifically Generalized ICP (GICP), Voxelized GICP and Distribution to Distribution Normal Distribution Transform (D2D NDT) are implemented in this work and can be used alternatively to our grid-based fitting approach. We use default parameters as defined by Fast-GICP.

\subsubsection{Grid-based fitting}
This work additionally introduces its own means of correlation estimation. For this, the radar point cloud $\mathcal{P}_{1}$ of pose $\mathbf{x}_1$ is converted to a grid $\mathcal{G}_1$ with a grid size of 0.1\,m. For each radar point the grid is updated by adding the probability density function values of a multivariate normal distribution centered at this point with a constant covariance for every point of 0.05\,m². This is aimed to represent the uncertainty in measurement we expect from the radar points as well as partly account for the discrete nature of the grid as opposed to the much higher resolution positional value of the radar points.

For pose $\mathbf{x}_2$, its radar point cloud $\mathcal{P}_{2}$ is initialized in the frame of pose $\mathbf{x}_1$ based on their relation in global space. A maximum relative angular error $\epsilon_R$ of 1° and a maximum relative translational error of $\epsilon_L$ 2\,m both in $x$ and $y$ direction is assumed between the two vehicle poses based on the accuracy observed of the trajectories within the dataset. 

Since applying a rotation to a grid is associated with some degree of interpolation loss, rotations are applied before converting the point cloud to grid. With a step $\Delta\theta$ of 0.1\,° we iterate over the range $[-\epsilon_R, \epsilon_R]$ and rotate the radar point cloud $\mathcal{P}_{2}$ by this value $\theta$. For each iteration $\mathcal{P}_{2}$ is then converted to grid $\mathcal{G}_{2,r}$ as described for $\mathcal{G}_1$.

We then traverse the grid $x \in [-\epsilon_L, \epsilon_L]$ and $y \in [-\epsilon_L, \epsilon_L]$ with steps $\Delta x$ and $\Delta y$ of 0.1\,m. At each step we apply the offset \(x, y\) to $\mathcal{G}_2$ and calculate the correlation $\rho_{1,2}$ of $\mathcal{G}_1$ and $\mathcal{G}_{2,r}$ by multiplying cell-wise and calculating the sum over all cells:
\begin{equation}
\rho_{1,2}(r, x, y) = \sum_{i,j} \left(\mathcal{G}_{1,r,i,j} \circ \mathcal{G}_{2,r,i+x,j+y} \right)
\end{equation}
.

From this we derive a three dimensional correlation matrix with dimensions for rotational as well as offset in x and y direction.

\begin{equation}
\mathbf{C}= [\rho_{1,2}(r,x,y)],
\text{with } r \in [-\epsilon_R, \epsilon_R], \; x , y \in [-\epsilon_L, \epsilon_L]
\end{equation}

The optimum transformation $T_{1,2}$ between the vehicle poses is then derived from the angular and linear offset with the highest correlation $\arg\max_{r,x,y} \rho_{1,2}(r,x,y)$ between $\mathcal{P}_{1}$ and $\mathcal{P}_{2}$:
\begin{equation}
T_{1,2} = \arg\max_{r,x,y} \rho_{1,2}(r,x,y).
\end{equation}

To further assess the quality of this correlation, we employ the standard score $Z$, defined as follows:
\begin{equation}
    Z_{1,2} = \frac{C - \mu(C)}{\sigma(C)}.
\end{equation}
This includes a normalization and therefore allows for comparability between correlations of different pose pairs and also shows the degree to which $\arg\max_{r,x,y} \rho_{1,2}(r,x,y)$ actually stands out compared to other calculated correlations for the same pose pair.

This approach is robust regarding missing individual radar points as systematically observed when comparing radar returns of objects at different distances and angles to the object, since no matching between individual points is necessary.

Correlation calculations are implemented to be offloaded to GPU accelerators and multiple entries of the "what-to-compare" list can also be computed in parallel using multiple GPUs.

We can extend each entry of our "what-to-compare" list with corresponding calculated correlation as well as standard score $Z$.

\subsection{Pose-graph optimization}
\label{subsec:posegraph}
Based on this extensive set of transformations between vehicle poses and their respective initial global positioning, a pose-graph optimization can be computed that outputs improved vehicle poses and aims to overcome the localization error-prone properties of pure GNSS based vehicle poses.

We employ GTSAM~\cite{gtsam} as the base framework as a computationally efficient implementation for smoothing and mapping operations. We construct the problem as a non-linear factor graph. For this, we include all vehicle poses for the drives present in the dataset with their trajectory as their initial guess as well as the first set of constraints. As the noise model for these measurements we utilize the standard deviation in both angular and translational direction provided as part of the dataset for each pose and assume a Gaussian distribution.

As the second set of constraints we utilize our "what-to-compare" list of transformations as calculated in \cref{subsec:correlation} as edges between their respective vehicle poses. For our grid-based fitting approach we further utilize the respective standard score $Z$ in inverted form for our pseudo noise model, including a weighting constant to account for the unit-less nature of this score. We further utilize the Huber robust error model ~\cite{huber1964robust} for robustification. For the baseline approaches we utilize a constant noise model.

Lastly, we employ the Levenberg–Marquardt optimizer ~\cite{levenberg1944method}, ~\cite{marquardt1963algorithm} on the non-linear factor graph to compute the aligned vehicle poses.

\subsection{Occupancy map}
\label{subsec:occu}
The aligned vehicle poses can be used to generate an occupancy map. For this, the global positions of all radar points are calculated based on their aligned vehicle poses. The result is a global radar point cloud that can be evaluated as in \cref{subsec:pc_aggregation}

Additionally, it can be used to calculate an occupancy map. For this, all points are registered as a normalized 2D histogram. This results in a global grid with a grid size of 0.1m that is further refined by applying a sigmoid function to the shifted and scaled value of each cell. This allows for simple noise reduction and richer contrast when plotted for qualitative evaluation.

\subsection{Lane boundary map}
\label{subsec:lane}
The improved poses can furthermore be utilized in combination with the road marking and road boundary detections present in the dataset to generate a map as proposed by Immel et al. \cite{immel2023hd} and described in \cref{subsec:sd4mad}. 

For evaluation purposes and to provide a baseline solution, the entire pipeline as described in \cref{subsec:sd4mad} is utilized with only slight changes to accommodate the changed feature detection properties, which mostly differ in actual longitudinal length.

Furthermore, to take advantage of our radar aligned poses, the stepwise expectation-maximization pose alignment step of this pipeline can be replaced by our radar aligned poses. Further steps are executed as proposed by Immel et al~\cite{immel2023hd}.

\begin{figure*}[t]
\centering
\begin{tabular}{ccccc}
\includegraphics[width=0.18\textwidth]{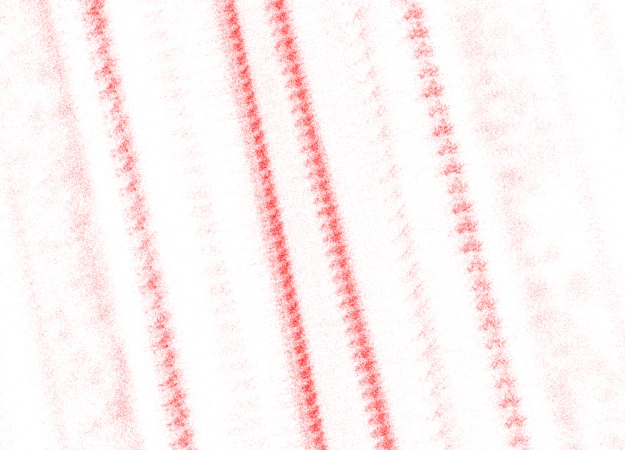} &
\includegraphics[width=0.18\textwidth]{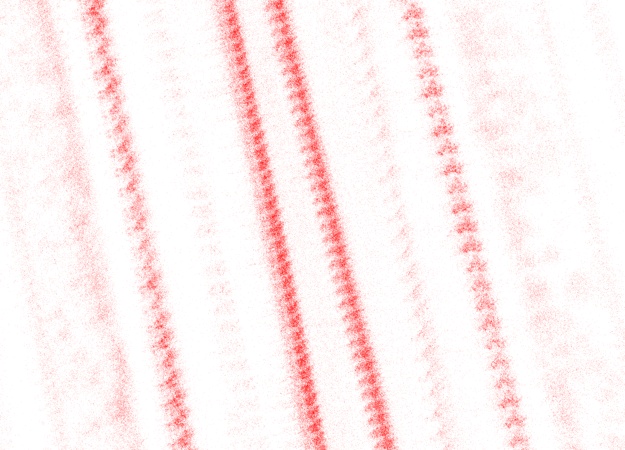} &
\includegraphics[width=0.18\textwidth]{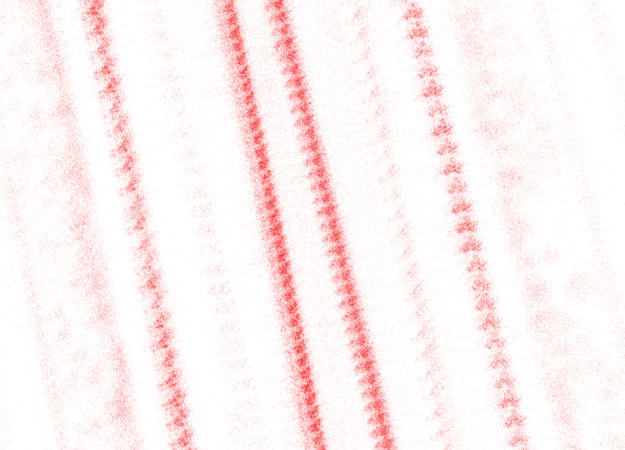} &
\includegraphics[width=0.18\textwidth]{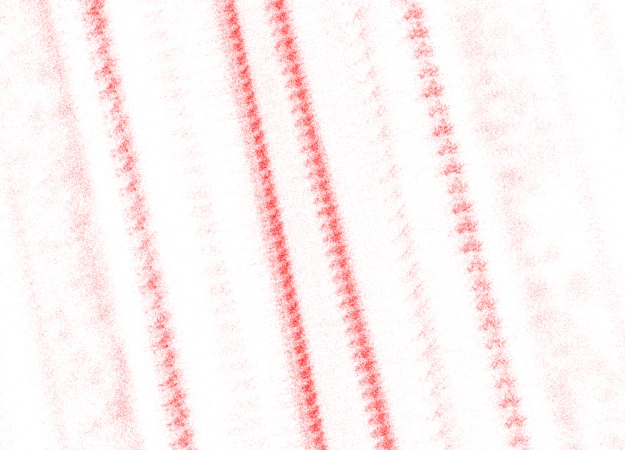} &
\includegraphics[width=0.18\textwidth]{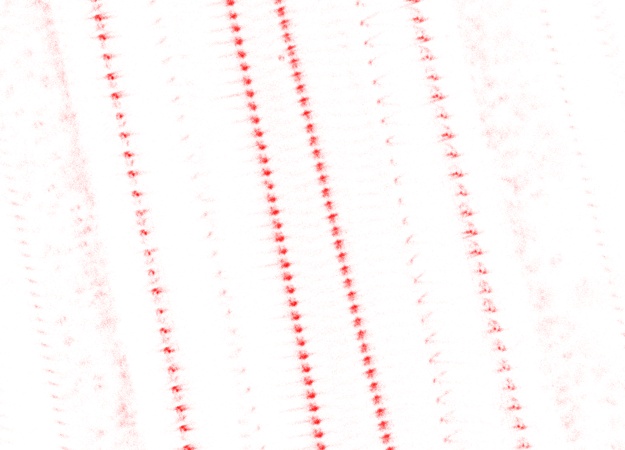} \\
Unaligned & GICP & VGICP & NDT & Ours \\
\end{tabular}
\caption{Occupancy maps of aligned point clouds. Only our grid-based fitting approach shows improvements over the unaligned point cloud, the others remain almost indistinguishable.}
\label{fig:mapeval_heatmaps}
\end{figure*}
\begin{figure*}[b]
    \centering
    \includegraphics[width=\textwidth]{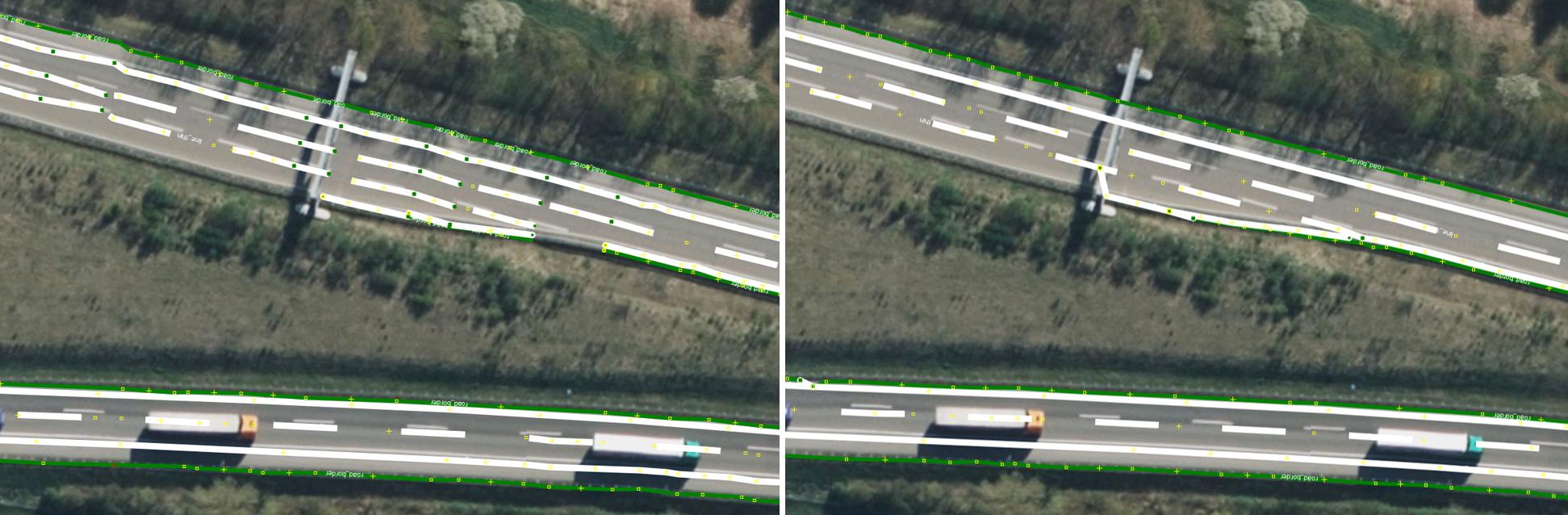}
    \caption{Qualitative comparison between expectation maximization based lane boundary aligned (left) and radar aligned (right) lane mapping results. In green the road boundary, in white solid and dashed lane dividers. Satellite imagery is not accurately georeferenced. The radar aligned map appears more smooth and shows less artifacts.}
    \label{fig:lanequali}
\end{figure*}
\begin{figure*}[t]
    \centering
    \begin{subfigure}[b]{0.45\linewidth}
        \centering
        \adjustbox{trim=0.6cm 0.6cm 1.6cm 1.4cm, width=\linewidth, clip}{
            \includesvg{images/error_over_meters_raw.svg}
        }
        \caption{Lane boundary aligned}
        \label{fig:raw_data_over_meters}
    \end{subfigure}
    \hfill
    \begin{subfigure}[b]{0.45\linewidth}
        \centering
        \adjustbox{trim=0.6cm 0.6cm 1.6cm 1.4cm, width=\linewidth, clip}{
            \includesvg{images/error_over_meters_radar.svg}
        }
        \caption{Radar Aligned}
        \label{fig:radar_data_over_meters}
    \end{subfigure}
    \caption{Offset  and non-offset error evaluated over the entire dataset. Whereas levels of offset and non-offset errors remain comparable, the fluctuation of errors of the lane boundary aligned approach greatly exceeds those of radar aligned poses}
    \label{fig:error_over_meters}
\end{figure*}
\begin{figure*}[bh!]
    \centering
    \begin{subfigure}[b]{0.45\linewidth}
        \centering
        \adjustbox{trim=0.6cm 0.6cm 1.6cm 1.4cm, width=\linewidth, clip}{
            \includesvg{images/boxplot_types_raw.svg}
        }
        \caption{Lane boundary aligned: The non-offset error as well as the general levels of the offset error are very comparable}
        \label{fig:boxplot_types_raw}
    \end{subfigure}
    \hfill
    \begin{subfigure}[b]{0.45\linewidth}
        \centering
        \adjustbox{trim=0.6cm 0.6cm 1.6cm 1.4cm, width=\linewidth, clip}{
            \includesvg{images/boxplot_types_radar.svg}
        }
        \caption{Radar Aligned: The non-offset error as well as the general levels of the offset error are very comparable}
        \label{fig:boxplot_types_radar}
    \end{subfigure}
    \caption{Average offset error and average non-offset error over the entire length of the labeled dataset}
    \label{fig:boxplots}
\end{figure*}
\section{Evaluation}
\subsection{Dataset}
\label{subsec:dataset}
For the evaluation of the proposed approaches, a vehicle fleet dataset is used consisting of a total of 60 drives along a german highway section with slightly varying start and end points. All drives combined cover a total distance of over 5000\,km. The drives cover both directions of this highway section over a greater length.

Poses include lateral and longitudinal coordinates as well as headings with lane-level accuracy. Driving speeds range from stopping to 36\,m/s.

For each pose of each drive, radar point clouds are provided with a 360\,° coverage of the surrounding, providing spacial information only.

Additionally, for each pose on-vehicle lane boundary detection are included. They provide further differentiation between the classes solid divider, dashed divider and road boundary. Each detection consists of a polyline in local coordinates, consisting of two to five points and a single digit length in meters.

For the purpose of quantitative evaluation, a section with a total length of 5.7\,km including both directions is hand-annotated on a lane marking level including road boundaries as well as further helpers necessary for the evaluation algorithm such as reference line and region of interest. This ground-truth is inherently limited due to its creation process of annotating in satellite imagery. While the overall road structure and relative accuracy between road elements can be well represented this way, absolute accuracy is limited due to the inaccurate georeferencing of satellite imagery. Additionally changes in road layout due to construction may introduce additional errors.

Quality of such a labeling strategy could be greatly improved with the usage of higher resolution, georeferenced aerial imagery in contrast to the satellite imagery utilized here, if its acquisition is feasible.

\subsection{Radar Mapping}
\subsubsection{Point cloud aggregation}
\label{subsec:pc_aggregation}
To compare the quality of the different correlation methods as introduced in \cref{subsec:correlation}, for each methods correlation output, the pose graph is computed according to \cref{subsec:posegraph} and the global radar point cloud calculated as in \cref{subsec:occu}. The same is done using unaligned poses as the global unaligned radar point cloud.

\begin{table}[h!]
\centering

\label{tab:mme}
    \begin{tabular}{l c c c c c}
    \hline
    \textbf{Metric} & \textbf{Unaligned} & \textbf{GICP} & \textbf{VGICP} & \textbf{NDT} & \textbf{Ours} \\
    \hline
    MME & -0.15507 & -0.150471 & -0.154754 & -0.15507 & \textbf{-0.490938} \\
    \hline
    \end{tabular}
    \caption{Mean Map Entropy on aligned point clouds. Lower is better. Only our grid-based fitting approach shows improvements over the unaligned point cloud.}
    \label{tab:mme}
\end{table}

Since no ground-truth is available, Mean Map Entropy (MME)~\cite{razlaw2015} is calculated for these global radar point clouds, specifically using the implementation of MapEval~\cite{hu2025mapeval}. The nearest neighbor search radius is set to 1m for this. \cref{tab:mme} shows the comparison between the different correlation methods. Whereas GICP, VGICP and D2D NDT do not provide significant improvements over the unaligned point cloud, our grid-based fitting approach does stand out.

\subsubsection{Occupancy maps}
\label{subsec:occupancy_maps}
\cref{fig:heatmap} shows radar occupancy maps overlaid on satellite imagery. On the left, the radar points are aggregated using unaltered vehicle poses. On the right, the poses are aligned as described in \cref{subsec:posegraph}. 

The most reoccurring features clearly visible as repeating local maxima in the occupancy grid in the right image are posts and pillars of the guardrails. On the unaligned side, there is no clear differentiation between individual posts; the middle divider section simply appears as one solid block.

Some artifacts do remain, however, after alignment, as visible on the bottom right of~\cref{fig:heatmap} as clear ghost radar returns. These appear to be some sort of reflection of the aforementioned middle divider section. As the structure of this middle section is still retained in the reflected instance, the reflection seems to yield repeatable results between drives and, whereas no part of the real world, can still facilitate localization in the map.

Radar occupancy maps are calculated for all correlation methods as introduced in \cref{subsec:correlation}. They are compared in \cref{fig:mapeval_heatmaps}. The scene again shows a highway scenario with both driving directions, each being lined by posts and pillars of the guardrails. Besides, similar artifacts as described for \cref{fig:heatmap} are visible. Only our grid-based fitting approach delivers a clear improvement here, with each post/pillar being clearly distinguishable from each other.

Both the quantitative (\cref{tab:mme}) as well as qualitative evaluation (\cref{fig:mapeval_heatmaps}) leads to the conclusion, that, for the data at hand, only our grid-based fitting approach can deliver noticeable improvements over the unaligned poses, all downstream tasks are therefore only computed based on this method.

\subsubsection{Lane boundary map}
The quality of the radar poses is further evaluated as part of the downstream tasks output of lane boundary map generation.

The lane boundary map output is generated based on the process described in \cref{subsec:lane}. Improved poses based on this work's approach are compared against the baseline of expectation maximization based lane boundary alignment as described in \cref{subsec:sd4mad}.

\cref{fig:lanequali} visualizes both approaches output. The absolute accuracy can only be roughly derived from the visualization due to the lack of high-accuracy georeferencing of the satellite imagery.

Whereas for the majority of the scene, both approaches generate comparable results with regards to the in driving direction lateral optimization, the longitudinal performance of this work's pose graph optimization based approach clearly results in an overall smoother output.

Sections, to which the mapping pipeline partly or fully fails to find a solution for one divider type or the entire scene are less common with the radar aligned approach as well. An example of this can be seen in the top left of \cref{fig:lanequali}, where there is a small missing section of dashed dividers observable with the baseline approach, whereas our approach delivers a continuous polyline output.

\subsection{Quantitative evaluation}
For the quantitative evaluation of the lane boundary map, the hand-annotated ground-truth as described in~\cref{subsec:dataset} is used.

\cref{subsec:sd4mad} introduces the two relevant evaluation metrics, for comparison visualized in \cref{fig:errors}. The metrics only evaluate lateral performance, the longitudinal performance can, however, be inferred based on the assumption of highway road features only rarely observing abrupt changes in longitudinal direction. Whereas lane merges, on and off ramps may cause some degree of abrupt change, for long stretches of roads, these should remain the exception.

\cref{fig:boxplots} visualizes the performance of the lane boundary aligned baseline approach on the left as well as this works radar aligned approach on the right as a box plot per detection class. For both the offset error as well as the non-offset error both approaches remain comparable over the entire range of labels.

With regards to the non-offset error, pure radar aligned poses therefore produce similar performance on a lane divider level compared to the specific optimization done to exactly these detections as part of the baseline approach. For both approaches, road boundary detections introduce a much greater error. This fits to the qualitative results as seen in~\cref{fig:lanequali}. Bigger fluctuations in road boundary detection can be attributed to the less pronounced feature expression compared to lane dividers.

With regards to the offset error, general levels provide limited meaningfulness due to the missing high accuracy georeferencing of the ground truth. \cref{fig:error_over_meters} visualizes offset and non-offset errors as an average over all classes plotted over the entire length of the annotated dataset. With the assumption of abrupt changes in lateral road layout to be uncommon, the strong fluctuations observed here shows the downside of the pure lateral optimization present in the baseline approach. The radar aligned approach relies on its pose graph optimization for both lateral and longitudinal optimization and delivers visibly smoother results here. This perfectly matches the qualitative results in~\cref{fig:lanequali}.
\FloatBarrier
\section{Conclusion}
 Based on fleet data we introduce a radar based pose optimization pipeline that allows for combined longitudinal and lateral alignment. The approach is based on robust grid based fitting and formulates the optimization as a pose graph, solving the problem for a globally aligned solution. Contrary to the baseline GICP, VGICP and D2D NDT approaches, correlation calculation remains robust and delivers clear improvements over the unaligned radar point clouds.
 
 We provide the option to generate a radar occupancy map based on the improved poses with qualitative results matching the expectations inferred from the overall road layout. Here we show contrast-rich feature clarity when visualized, focusing on omnipresent guardrail posts as the main feature type observable in the map.
 
 Furthermore, we utilize these improved poses as part of a lane boundary-based HD mapping pipeline, improving overall performance and overcoming the original approach's limitations with regard to longitudinal optimization. Lateral fluctuations, observable both in the qualitative results as well as the plotted offset error, can be effectively mitigated with this approach.

Further work could focus on improving other aspects of the pipeline, like the polyline graph generation portion. Furthermore, since the current approach is only evaluated for highway scenarios, performance in urban environment is still untested. The pipeline could be adapted and evaluated on such scenarios.
 
% \small
%\section*{Acknowledgments}

\printbibliography

\end{document}